\pdfoutput=1
\documentclass[11pt]{article}
\usepackage[final]{acl}
\usepackage{times}
\usepackage{latexsym}
\usepackage[T1]{fontenc}
\usepackage[utf8]{inputenc}
\usepackage{microtype}
\usepackage{inconsolata}
\usepackage{graphicx}
\usepackage{booktabs}
\usepackage{tabularx} 
\usepackage{makecell}
\usepackage[table]{xcolor}   
\usepackage{wrapfig}
\usepackage{threeparttable}
\usepackage{amsmath}
\usepackage{amsfonts}
\usepackage{bm}
\usepackage{textcomp}
\usepackage{xcolor}
\usepackage{colortbl}
\usepackage{tcolorbox}
\usepackage{color}
\usepackage{enumitem}
\usepackage{multirow}
\usepackage{geometry}
\usepackage[ruled,vlined]{algorithm2e}
\usepackage{hyperref}
\usepackage{xcolor} 
\definecolor{darkgreen}{rgb}{0.0, 0.5, 0.0}

\usepackage{booktabs}
\usepackage{pifont}
\newcommand{\xmark}{\ding{55}}

\usepackage{cleveref}
\crefformat{section}{\S#2#1#3}
\crefformat{subsection}{\S#2#1#3}
\crefformat{subsubsection}{\S#2#1#3}
\crefrangeformat{section}{\S\S#3#1#4 to~#5#2#6}
\crefmultiformat{section}{\S\S#2#1#3}{ and~#2#1#3}{, #2#1#3}{ and~#2#1#3}
\crefmultiformat{subsection}{\S\S#2#1#3}{ and~#2#1#3}{, #2#1#3}{ and~#2#1#3}
\Crefformat{figure}{#2Fig.~#1#3}
\Crefmultiformat{figure}{Figs.~#2#1#3}{ and~#2#1#3}{, #2#1#3}{ and~#2#1#3}
\Crefformat{table}{#2Tab.~#1#3}
\Crefmultiformat{table}{Tabs.~#2#1#3}{ and~#2#1#3}{, #2#1#3}{ and~#2#1#3}
\Crefformat{appendix}{Appx.#2#1#3}
\crefmultiformat{appendix}{Appx.#2#1#3}{ and~#2#1#3}{, #2#1#3}{ and~#2#1#3}
\crefformat{algorithm}{Alg.~#2#1#3}
\Crefformat{equation}{Eq.~#2#1#3}
\newcommand{\myparagraph}[1]{\vspace{0.3em}\noindent{{\bf #1.}}}
\newcommand{\modelname}{\textsc{Socrates}\xspace} 
\newcommand{\modelnameQW}{\textsc{Socrates-Qwen-14B}\xspace} 
\newcommand{\modelnameLA}{\textsc{Socrates-LLaMA-8B}\xspace} 
\newcommand{\dataname}{\textsc{SocSci210}\xspace} 
\definecolor{defaultcolor}{gray}{0.9}

\title{Finetuning LLMs for Human Behavior Prediction in Social Science Experiments}

\author{
 \textbf{Akaash Kolluri\textsuperscript{1*}},
 \textbf{Shengguang Wu\textsuperscript{1*}},
 \textbf{Joon Sung Park\textsuperscript{1}},
 \textbf{Michael S. Bernstein\textsuperscript{1}}
\\
\\
 \textsuperscript{1}Stanford University
\\
 \texttt{
  \{akaash,shgwu\}@stanford.edu}
}

\begin{document}
\maketitle
\begin{abstract}

Large language models (LLMs) offer a powerful opportunity to simulate the results of social science experiments. In this work, we demonstrate that finetuning LLMs directly on individual-level responses from past experiments meaningfully improves the accuracy of such simulations across diverse social science domains. We construct \dataname via an automatic pipeline, a dataset comprising 2.9 million responses from 400,491 participants in 210 open-source social science experiments. Through finetuning, we achieve multiple levels of generalization. In completely unseen studies, our strongest model, \modelnameQW, produces predictions that are 26\% more aligned with distributions of human responses to diverse outcome questions under varying conditions relative to its base model (\texttt{Qwen2.5-14B}), outperforming \texttt{GPT-4o} by 13\%. By finetuning on a subset of conditions in a study, generalization to new unseen conditions is particularly robust, improving by 71\%.  Since \dataname contains rich demographic information, we reduce demographic parity difference, a measure of bias, by 10.6\% through finetuning. Because social sciences routinely generate rich, topic-specific datasets, our findings indicate that finetuning on such data could enable more accurate simulations for experimental hypothesis screening. We release our data, models and finetuning code at \url{stanfordhci.github.io/socrates}.

\end{abstract}

\section{Introduction} \label{sec:intro}

\begingroup
\renewcommand\thefootnote{}
\footnotetext{* Denotes equal contribution.}
\addtocounter{footnote}{0}
\endgroup

\begin{figure}[h!]
    \centering
     \includegraphics[width=1\linewidth]{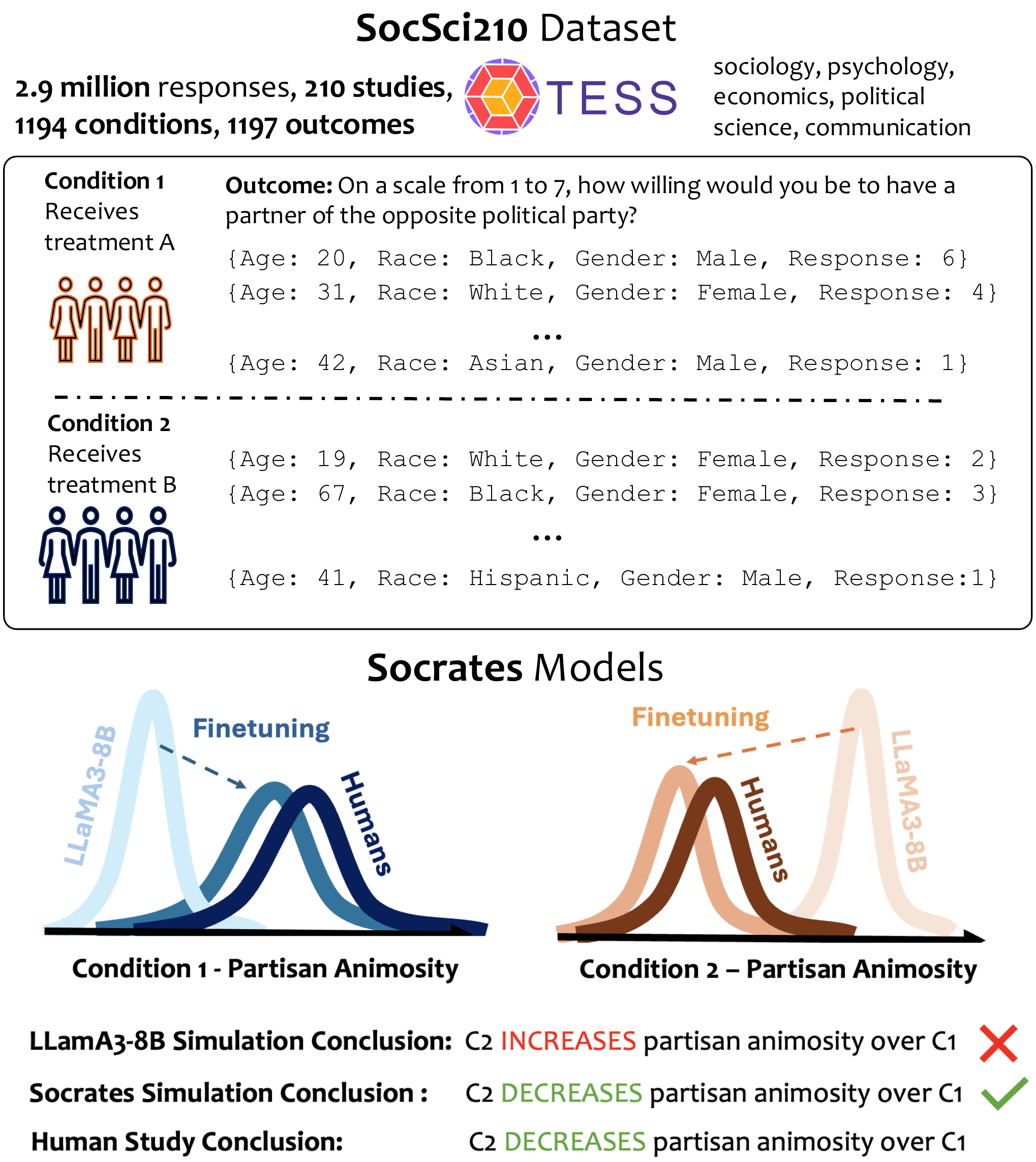}
    \caption{We release \dataname, a large-scale dataset built from open-source social science experiments. Through finetuning, we create behavioral prediction models \modelnameLA and \modelnameQW, which predict responses that are 12.1\% and 13.2\% respectively more aligned with human response distributions to outcomes under diverse experimental conditions, relative to \texttt{GPT-4o}.}
    \label{fig:teaser}
\end{figure}

Large language models have shown impressive potential to simulate human behavior \cite{park2024generative, hewitt2024predicting, kim2023ai}. For social science experiments, simulations enable researchers to screen and iterate on hypotheses before committing to costly studies \cite{rothschild2024opportunities, hewitt2024predicting, wang2025chatgpt}. Accordingly, LLM-based simulation methods have been explored across various social science disciplines \cite{argyle2023out, horton2023large, brand2023using}.

Previous work simulating human responses has used direct prompting such as with demographic personas \cite{hewitt2024predicting}, human conversations \cite{cho2024llm}, and detailed life narratives \cite{park2024generative, moon2024virtual}. Still, LLMs routinely distort opinion distributions \cite{bisbee2024synthetic, gao2024take}, overestimate effect sizes in experimental manipulations by \textbf{2 to 10} times \cite{park2024generative, hewitt2024predicting}, and incorrectly predict significant effect directions \textbf{10 - 32\% of the time} \cite{hewitt2024predicting, bisbee2024synthetic}. LLMs further introduce biases that flatten variation across demographic groups \cite{wang2025large}. These error cases currently limit the viability of effective LLM-simulations for social science experiments.

Recent work has begun exploring the viability of finetuning language models for improved human response prediction \cite{suh2025language, chu2023language, lu2025beyond, binz2024centaur} and has demonstrated generalization on their specific tasks (\emph{e.g.}, cognitive science, public opinion). 

In this work, we broaden the domain and scope of prior fine-tuning work in pursuit of a general purpose, domain-agnostic human behavior prediction model. To enable this, we first construct \dataname, a standardized, large-scale dataset comprising \textbf{2.9 million} individual responses from over \textbf{400,000} participants across \textbf{210} social science studies. All studies were drawn from NSF's Time-sharing Experiments for the Social Sciences—peer-reviewed, high-powered experiments spanning multiple disciplines (\emph{e.g.}, economics, political science, behavioral psychology) and conducted on nationally representative samples with rich demographic reporting \cite{tess}. We design an LLM agent to convert each study’s data into a consistent text-based representation describing respondent demographic profiles, the experimental questions, and the recorded responses.

Using this dataset, we provide a comprehensive comparison of finetuning methods (supervised fine-tuning, augmenting with reasoning traces, contrastive preference optimization) against various prompting baselines (reasoning and in-context learning) on both proprietary \texttt{GPT-4o} and open-source \texttt{LLaMa3-8B}~\citep{llama3}, \texttt{Qwen2.5-14B}~\citep{qwen2.5} LLMs. Through our evaluations, we highlight that supervised fine-tuning greatly improves distributional alignment between predicted responses and human responses, while contrastive preference optimization leads to the best prediction accuracy for individual responses. Notably, relative to their base models, fine-tuning improves alignment with human response distributions in unseen studies by 30\% for \texttt{LLaMa3-8B} and 26\% for \texttt{Qwen2.5-14B}. We further demonstrate robust generalization to unseen participants, conditions and outcomes. Finally, we highlight finetuning improve demographic bias in predictions by 10+\%.


Our main contributions are:

1. We release \dataname, a standardized, large-scale dataset comprising 2.9 million individual responses from over 400,000 participants ($5\times$ the number participants of prior work's datasets) with rich demographic reporting across 210 social-science studies spanning multiple disciplines. 

2. We present \modelnameLA and \modelnameQW finetuned on \dataname, which, relative to \texttt{GPT-4o}, generate predictions that align 12.1\% and 13.2\% better to human response distributions, reflecting 26+\% performances gains relative to their base models.

3. Motivated by practical use cases of social scientists with in-domain data, we demonstrate robust generalization at various levels. Finetuning on as little as 10\% of an experiment's data reduces prediction error by 13\% on unseen participants, and training on subsets of experimental conditions or outcomes boosts generalization to unseen conditions by 71\% and unseen outcomes by 49\%.

\section{Related Work}

\begin{table*}[ht]
\centering
\footnotesize
\begin{tabular}{@{} l rrr ll l @{}}
\toprule
\multirow{2}{*}{\textbf{Dataset}}
  & \multicolumn{3}{c}{\textbf{Size}}
  & \multicolumn{2}{c}{\textbf{Feats.}}
  & \multirow{2}{*}{\textbf{Domain}} \\
\cmidrule(lr){2-4} \cmidrule(lr){5-6}
  & \textbf{Source} & \textbf{Individuals} & \textbf{Total Data Points}
  & \textbf{D} & \textbf{I} & \\
\midrule
\texttt{Psych-101} \cite{binz2024centaur}       & 160 experiments   & 60,000   & 10,000,000& \xmark & \checkmark & Psychology            \\
\texttt{SubPOP} \cite{orlikowski2025beyond}           & 3,362 questions  & —         & 70,000          & \checkmark & \xmark & Public Opinion        \\
\texttt{E-commerce} \cite{lu2025beyond}   & 31,865 sessions  & 3,526    & 230,965        & \xmark & \checkmark & E-commerce \\
\texttt{OpinionQA} \cite{pmlr-v202-santurkar23a}  & $\sim$1,500 questions & — & ~90,000    & \checkmark & \xmark & Public Opinion     \\
\multirow{3}{*}{\texttt{Be.FM} \cite{xie2025fm}} 
  & 50 questions       & 17,667 & 883,350   & \checkmark & \checkmark & Big-5 Personality \\
  & 6 games & 68,790 & 82,057  & \checkmark & \checkmark & Behavioral Econ. \\
  & 2,703 abstracts & — &  2,703 & \xmark & \xmark & Behavioral Science \\

\midrule
\dataname\ (Ours)             & 210 experiments   & \textbf{400,491}  & 2,900,000   & \textbf{\checkmark} & \textbf{\checkmark} & Social Sciences \\
\bottomrule
\end{tabular}
\caption{Comparison of our \dataname to dataset characteristics used in prior finetuning work. Under ``Feats (features)'' column, ``D'' indicates if the dataset includes participant demographics, and ``I'' refers to if training samples are done at the individual level (as opposed to the aggregate distribution level). When training occurs at the aggregate level, we omit the total number of individuals used to construct the dataset as they do not maintain individual granularity.  }
\label{tab:dataset-comparison}
\end{table*}

\myparagraph{Datasets for Human Response Finetuning}
Recent works have assembled large-scale public-opinion datasets and used them to finetune LLMs. For instance, \citet{pmlr-v202-santurkar23a} train on opinion distributions from 60 U.S. demographic groups over 500 contentious questions. Likewise, \citet{suh2025language} compile 3,362 survey questions with the responses distributions from 70,000 demographics, demonstrating a great breadth of topic diversity.

Other works have focused on datasets of individual decision-making across different behavioral contexts. \citet{binz2024centaur}, for example, introduce Psych-101, which contains over 10 million choices from 60,000 participants across 160 cognitive-science experiments. \citet{orlikowski2025beyond} collect and finetune on 60,000 individuals reactions to different texts to explore how sociodemographic factors shape perception, and \citet{lu2025beyond} finetune on 230,965 logged decisions from 3,526 users to predict web-action generation. \citet{zhu2025using} explores reinforcement fine-tuning for reasoning trace generation from a dataset of ~13,000 risky human choices produced by \citet{peterson2021using}. \citet{xie2025fm} combines multiple types of datasets for fine-tuning, aggregating data from 17,667 individual surveys and economic games played by 68,790 individuals. (they also include titles and abstracts from 2,703 behavioral science publications). Although rich in behavioral detail, these datasets do not yet capture the full diversity of social science disciplines that simulations could enable at a granular, individual level.

Our work attempts to bridge these gaps by constructing a dataset that expands upon the coverage of human behavioral sciences covered in prior datasets while enabling granular, individual-level responses in behavioral contexts. Increased scientific topic breadth is a key enabler of engineering a shared agent model that can enable any simulation across any social science context.

\begin{figure}[t]
    \centering
    \includegraphics[width=1\linewidth]{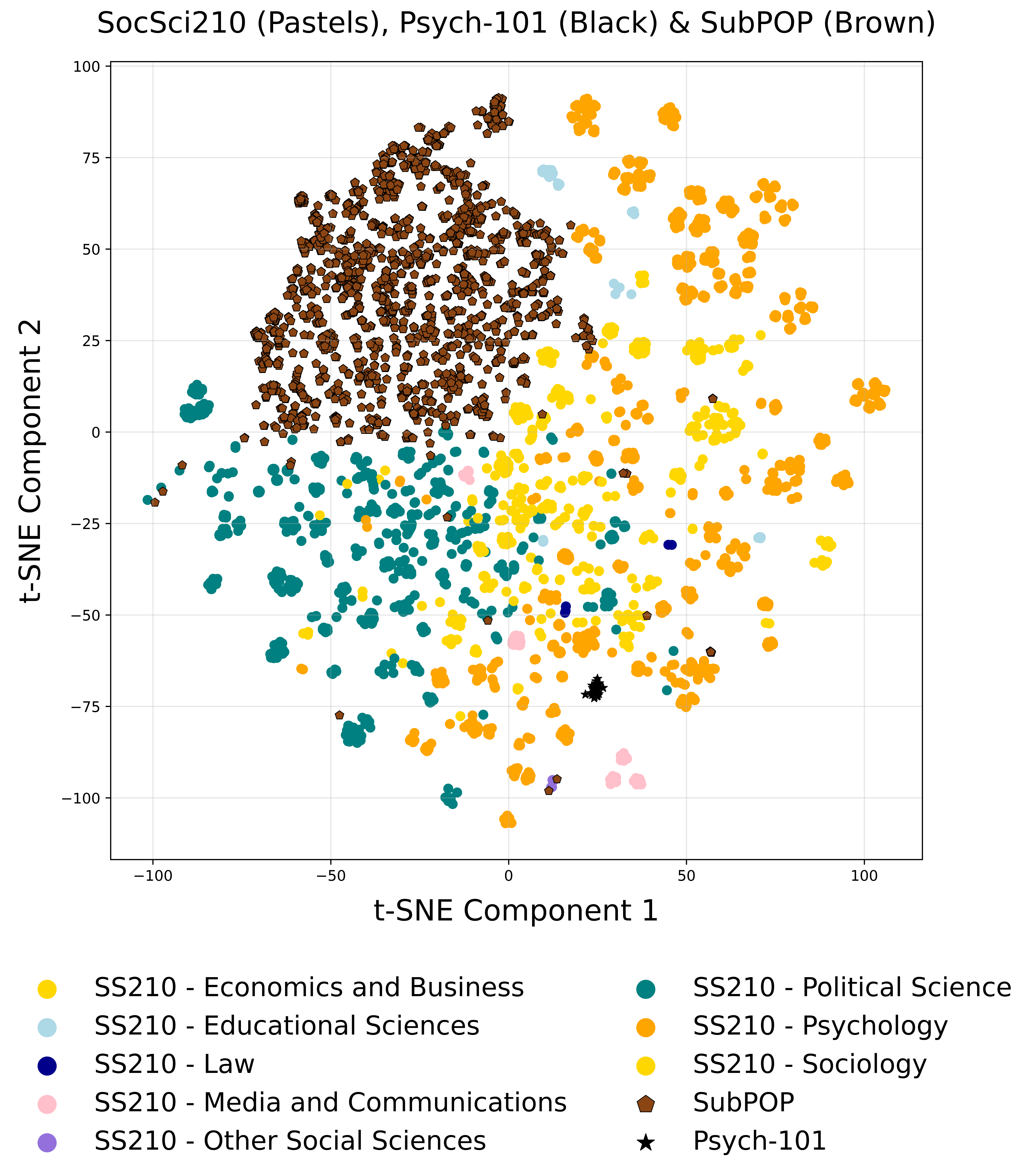}
    \caption{t-SNE projected embedding space of questions in \dataname, compared to \texttt{SubPop} \cite{suh2025language} and \text{Psych101} \cite{binz2024centaur}. \dataname shows much broader topic diversity across social science disciplines.}
    \label{fig:dataembeddings}
\end{figure}

\myparagraph{LLM Finetuning Methods}
Finetuning adapts pretrained LLMs for specific tasks, such as following user instructions~\citep{instructgpt,wang2022self,zhang2023instruction} or learning social skills~\citep{liu2023training,yang2024social,wu2024rethinking}. Apart from supervised finetuning (SFT), reinforcement learning from human feedback (RLHF) \citep{christiano2017deep,ziegler2019fine}, as well as the simplified DPO~\citep{rafailov2024direct} and SimPO~\citep{meng2024simpo}, all use paired data of \texttt{\{preferred response, dispreferred response\}} for finetuning, which aligns model outputs to annotated human preferences.
More recently, reasoning models~\citep{deepseek,abdin2025phi,qwq32b} trained with RL show strong performance in improving LLM capability by exploring chain-of-thought~\citep{cot}, especially for solving complex tasks. The reasoning abilities of larger teacher models can also be distilled into smaller models via finetuning on the teacher-generated reasoning traces~\citep{zhao20251,open_thoughts_2025}.

In this work, we provide a comparative evaluation of SFT and DPO—contrasting the scenarios in which each is most effective (see \Cref{sec:finetuning}). Prior finetuning studies in this domain have explored direct SFT \cite{suh2025language, binz2024centaur}, SFT augmented with reasoning \cite{lu2025beyond}, and reinforcement fine-tuning with GRPO \cite{zhu2025using}.

\begin{figure*}[tb]
    \centering
    \includegraphics[width=1.0\linewidth]{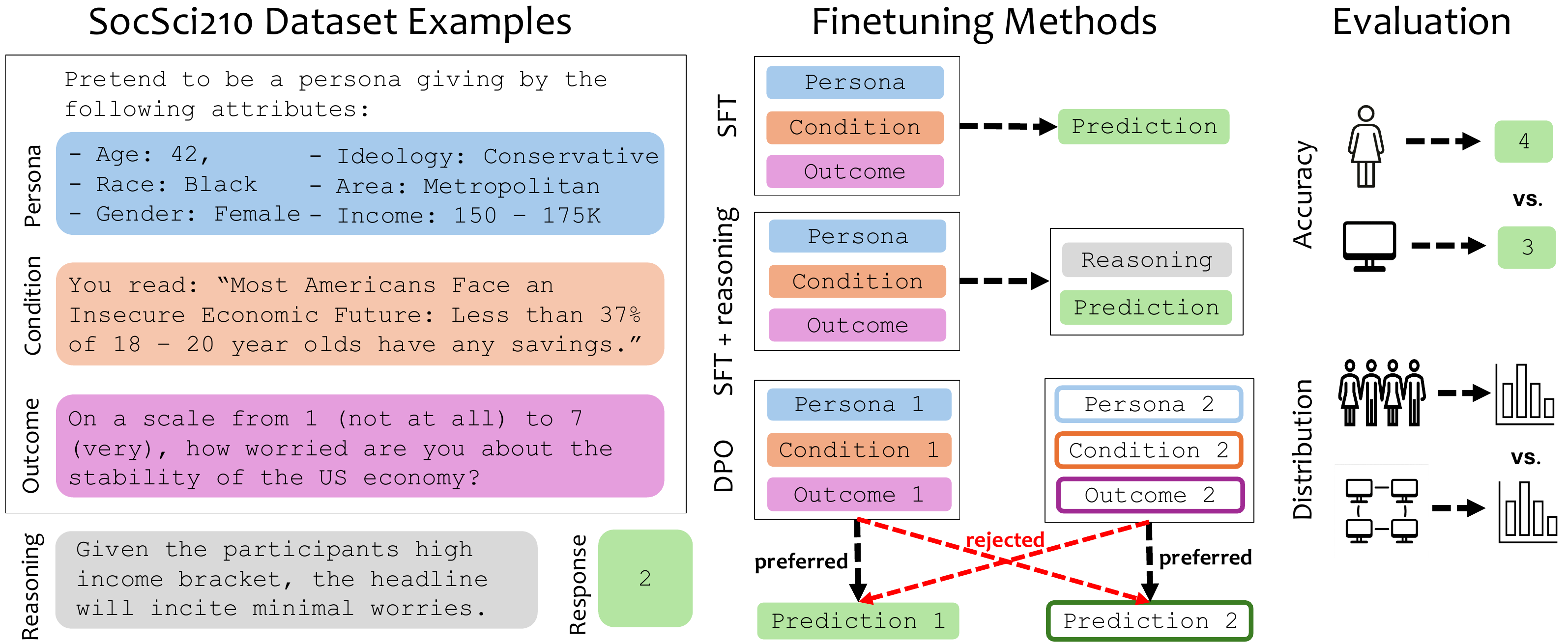}
    \caption{Overview of our task formulation, methods, and evaluation. Our dataset contains information on personas, conditions, outcomes, and predictions. We compare SFT, SFT on reasoning traces, and DPO. Our evaluation measures performance gains on both predicting individual accuracy and aggregate distributions under conditions.}
    \label{fig:methods-fig}
\end{figure*}

\section{Task Formulation} \label{sec:task_formulation}

\subsection{Task Description} \label{subsec:task_desc}

We finetune an LLM to predict the responses of an individual (of some demographic) to a stimulus in an experiment with multiple treatments. \Cref{fig:methods-fig} shows an example prediction. We consider questions that are ordinal (\emph{e.g.}, ``On a scale from 1 to 7, how satisfied are you with your life?'') or binary (\emph{e.g.}, ``Would you buy this? Answer yes or no.''). In the dataset of experiments we draw from, outcomes were primarily ordinal or binary; restricting to only these response types standardized our model evaluations, with minimal data size reduction.

Formally, in our dataset $D$, a person is represented as $P$, a set of unique attributes. In our dataset, these attributes are demographic characteristics. Each experiment is set up as $E =([c_1,...,c_j],[o_1, ..o_i])$ with $j$ different conditions and $i$ outcome questions. The goal of the experiment is to see how different conditions impact participant responses to a given outcome. Studies may have a between-subject design (i.e., participants are randomized into a single condition and answer multiple outcome questions (between-subject studies) or they may have an within-subject design (participants are randomized into multiple conditions and answer the same outcome). In either setup, each participant gives $k$ responses, $r$, each corresponding to a question representing a $(c,o)$ pair, which we refer to as the stimuli. Note, the randomization process occurred when the study of the original dataset was conducted, so we do not randomize the participants into the conditions, we simply tag the participants' responses with the condition they were in in the original study. For every individual participant in a study, we have $k$ tuples of $(P, c, o, r)$. We finetune an LLM, $F'$, to learn $F(P,c,o) \implies r$.

\subsection{Evaluation} \label{sec:eval_metrics}

We evaluate the performance of our finetuned model $F’$ at two levels.

\myparagraph{Individual Response Accuracy} 
Given the ordinal property of all predictions (\Cref{subsec:task_desc}), we compute a normalized accuracy between predicted and actual responses as
$$\text{Acc.} = 1 - \frac{1}{N} \sum_{(P, c, o)}  \frac{\bigl|F'(P,c,o) - r\bigr|}{r_{\max}-r_{\min}}$$
where $r_{\max}, r_{\min}$ represent the maximum and minimum value of $r$ under a specific condition and outcome in the ground truth responses (\emph{i.e.}, the bounds of the response scale). This value is computed for each study, and then averaged across all studies to get a final individual response accuracy score. 

\myparagraph{Distribution Alignment Under Conditions} 
While predicting individual responses is important, accuracy is upper-bounded by the non-deterministic nature of $F(P,c,o)$ (\emph{i.e.}, the same demographic may have different responses to the same stimuli, so $F(P,c,o)$ is a distribution). The primary goal of social-science experiments is to compute an end statistic that represents how responses to outcome questions vary across stimulus conditions. The exact statistical analyses in these contexts (\emph{e.g.}, t-tests, ANOVA, regressions) depend on the type of experiment being conducted \cite{maravelakis2019use}, but all statistics depend on the underlying distribution of the responses to outcome questions under each condition. Because of this, distributional alignment for this context is more important than measuring just accuracy (we provide further examples and justification of this in \cref{appendix:accuracy-vs-alignment}). Thus, we also measure whether the distribution of responses to each outcome under each experimental condition aligns with corresponding human responses. 

Specifically, for each (condition, outcome) pair $(c,o) \in E$, we compare the distribution of predicted responses across all participants assigned to $c$ against the empirical distribution of actual responses. Similar to \citet{suh2025language}, we compute distributional alignment using the Wasserstein distance, which estimates both the shape and mean of distributions. For consistency, we first standardize all distributions to be between $[0,1]$ by subtracting $r_{\min}$ and dividing by $r_{\max} - r_{\min}$. We average the Wasserstein distance across all condition outcomes pairs in a study to get an aggregate score for the study. We then average each study's score across all studies to get the final score. A lower Wasserstein distance indicates better alignment between predicted and empirical response distributions.

\subsection{Dataset Construction}

\myparagraph{Data Source} We collect studies from NSF's Time-sharing Experiments for the Social Science (TESS) project \cite{tess}, a repository of peer-reviewed experiments across various social disciplines (\emph{e.g.}, psychology, political science, and economics). TESS studies are nationally representative and high powered (studies in \dataname have mean 1907 and median 1954.5 participants).

\myparagraph{Reconstruction} 
We employ a data‐construction agent -- powered by OpenAI's \texttt{o4-mini-high} -- to automatically parse the source data into \texttt{\{persona, stimuli, response\}} formats from each study, where the stimuli is tagged by its respective condition and outcome question. We detail the reconstruction workflow in \Cref{sec:data_reconstruction}. 
In total, our agent successfully reconstructs 210 studies.

\myparagraph{Final Dataset Statistics}
\dataname comprises 2.9 million individual responses spanning 1197 outcomes and 1194 conditions (yielding collectively 5,998 unique stimuli) from 400,491 participants. 
Our \dataname includes responses from \textbf{five times as many individuals as prior finetuning work}. \Cref{tab:dataset-comparison} offers an explicit comparison with prior datasets used for human behavior finetuning.
Apart from being large in scale, \dataname is also diverse across disciplines.
\Cref{fig:dataembeddings} shows the embedding space of our stimuli with other large-scale fine-tuning datasets \cite{binz2024centaur, suh2025language}, illustrating this broad topic diversity.

\section{Finetuning Methods}  \label{sec:finetuning}

To finetune LLMs for simulating responses, we experiment with supervised finetuning (SFT), SFT on oracle reasoning traces, and contrastive preference tuning via DPO~\citep{rafailov2024direct}.

\myparagraph{Supervised Finetuning (SFT)}
Given our dataset \(\mathcal{D}\), an individual persona \(P\), experiment condition \(c\), and outcome question \(o\), we form a prompt \(q\) that asks the model to predict the individual’s response (see \Cref{sec:prompt} for the template).  
Let \(F'\) denote the model being finetuned. The SFT objective minimizes the negative log-likelihood (cross-entropy) of the ground-truth response \(r\):
\begin{equation}
\mathcal{L}_{\text{SFT}}\!\bigl(F'\bigr)
= -\,\mathbb{E}_{(q,r)\sim\mathcal{D}}\!\left[\,
\log F'(r \mid q)\,\right]
\end{equation}

\myparagraph{Augmentation with Oracle Reasoning Traces}
Prior work has used explanatory reasoning for predicting human behavior~\citep{park2024generative} and improving fine-tuning performance \citep{lu2025beyond}, so we also augment SFT with oracle reasoning.
Specifically, given the LLM prompt and the corresponding human response, we query \texttt{GPT-4o-mini} to generate reasoning traces explaining the human decision from a social scientist's perspective (see \Cref{sec:prompt}). These oracle-generated reasoning traces are incorporated into the target output, enriching the response $r$ with explicit rationales\footnote{While we use \texttt{GPT-4o} for our main evaluation, we opted to generate the oracle reasoning traces with \texttt{GPT-4o-mini} due to practical cost constraints. Since these traces are produced oracle-style, we believe any difference in reasoning quality between the models is minimal.}.

\myparagraph{Contrastive Finetuning via Preference Optimization}
To enhance the model's ability to differentiate responses based on variations in conditions $c$, demographics $P$, or outcome questions $o$, we construct paired data by varying these components and contrasting the corresponding responses.

For demographic contrastive pairs (see \Cref{subsec:gen-participant}), we take a focal persona \(p_{pos}\) and randomly sample a contrasting persona \(p_{neg}\) from the dataset \(\mathcal{D}\) under the same condition \(c\) and outcome question \(o\), while ensuring their responses differ (\(r_{neg}\)).  
Each pair specifies that \((p_{pos}, c, o, r_{pos})\) is preferred over \((p_{pos}, c, o, r_{neg})\).  
Following \citet{rafailov2024direct}, the DPO objective is

\begin{multline}
\mathcal{L}_{\text{DPO}}\!\bigl(F';F\bigr)=\\
-\,\mathbb{E}_{(q,r_{pos},r_{neg}) \sim \mathcal{D}}
\Bigl[
\log \sigma\!\Bigl(
\beta \log
\frac{F'(r_{pos}\mid q)}{F(r_{pos}\mid q)}\\
-
\beta \log
\frac{F'(r_{neg}\mid q)}{F(r_{neg}\mid q)}
\Bigr)
\Bigr],
\end{multline}

where \(q\) concatenates \(p_{pos}\), \(c\), and \(o\) into a single prompt (see \Cref{sec:prompt}); \(F'\) is the finetuned model, \(F\) the fixed reference model, \(\sigma\) is the sigmoid function, and \(\beta\) scales the preference impact.


\section{Experiments}

\subsection{Training Configurations}
We train \texttt{LLaMA3-8B-Instruct}~\citep{llama3} and \texttt{Qwen2.5-14B-Instruct}~\citep{qwen2.5} as representative base LLMs of different sizes.  \Cref{sec:implement_details} has complete training details.

\subsection{Baselines}

\begin{table*}[t]
\footnotesize
\centering
 \newcolumntype{Y}{>{\centering\arraybackslash}p{1.6cm}} 
 \newcolumntype{Z}{>{\centering\arraybackslash}p{1.6cm}} 
\begin{tabular}{p{3.5cm}  YZZ YZZ}
\toprule
\multirow{2}{*}{\textbf{Model Variant}} &
\multicolumn{3}{c}{\textbf{Accuracy $\uparrow$}} &
\multicolumn{3}{c}{\textbf{Distribution $\downarrow$}} \\
\cmidrule(lr){2-4} \cmidrule(lr){5-7}
& \textbf{Score} & \textbf{$\% \Delta$ vs Base} & \textbf{vs \texttt{GPT-4o}}
& \textbf{Score} & \textbf{$\% \Delta$ vs Base} & \textbf{vs \texttt{GPT-4o}} \\
\midrule
\rowcolor{gray!12}
\multicolumn{7}{l}{\textbf{Proprietary Models}} \\
GPT-4o Base & 72.9 & -- & -- & 0.174 & -- & -- \\
\quad + Few-shot (5) & 73.2 & \textcolor{darkgreen}{0.4\%} & \textcolor{darkgreen}{0.4\%} & 0.161 & \textcolor{darkgreen}{7.5\%} & \textcolor{darkgreen}{7.5\%} \\
\quad + Reasoning & 73.1 & \textcolor{darkgreen}{0.3\%} & \textcolor{darkgreen}{0.3\%} & 0.169 & \textcolor{darkgreen}{2.9\%} & \textcolor{darkgreen}{2.9\%} \\
\rowcolor{gray!12}
\multicolumn{7}{l}{\textbf{Open-Source Models}} \\
LLaMA3-8B Base & \underline{70.3} & -- & \textcolor{red}{-3.6\%} & 0.219 & -- & \textcolor{red}{-25.9\%} \\
\quad + Few-shot (5) & 68.9 & \textcolor{red}{-2.0\%} & \textcolor{red}{-5.5\%} & 0.212 & \textcolor{darkgreen}{3.2\%} & \textcolor{red}{-21.8\%} \\
\quad + Reasoning & 69.8 & \textcolor{red}{-0.7\%} & \textcolor{red}{-4.3\%} & 0.174 & \textcolor{darkgreen}{20.6\%} & -- \\
\quad + SFT & 69.1 & \textcolor{red}{-1.7\%} & \textcolor{red}{-5.2\%} & \textbf{0.153} & \textcolor{darkgreen}{30.1\%} & \textcolor{darkgreen}{12.1\%} \\
\quad + SFT w/ Reasoning & 67.5 & \textcolor{red}{-4.0\%} & \textcolor{red}{-7.4\%} & \underline{0.165} & \textcolor{darkgreen}{24.7\%} & \textcolor{darkgreen}{5.2\%} \\
\quad + DPO & \textbf{72.6} & \textcolor{darkgreen}{3.3\%} & \textcolor{red}{-0.4\%} & 0.185 & \textcolor{darkgreen}{15.5\%} & \textcolor{red}{-6.3\%} \\
Qwen2.5-14B Base & \underline{72.9} & -- & -- & 0.205 & -- & \textcolor{red}{-17.8\%} \\
\quad + Few-shot (5) & 71.9 & \textcolor{red}{-1.4\%} & \textcolor{red}{-1.4\%} & 0.196 & \textcolor{darkgreen}{4.4\%} & \textcolor{red}{-12.6\%} \\
\quad + Reasoning & 72.7 & \textcolor{red}{-0.3\%} & \textcolor{red}{-0.3\%} & 0.166 & \textcolor{darkgreen}{19.0\%} & \textcolor{darkgreen}{4.6\%} \\
\quad + SFT & 69.5 & \textcolor{red}{-4.7\%} & \textcolor{red}{-4.7\%} & \textbf{0.151} & \textcolor{darkgreen}{26.3\%} & \textcolor{darkgreen}{13.2\%} \\
\quad + SFT w/ Reasoning & 67.6 & \textcolor{red}{-7.3\%} & \textcolor{red}{-7.3\%} & \underline{0.164} & \textcolor{darkgreen}{20.0\%} & \textcolor{darkgreen}{5.7\%} \\
\quad + DPO & \textbf{74.0} & \textcolor{darkgreen}{1.4\%} & \textcolor{darkgreen}{1.4\%} & 0.181 & \textcolor{darkgreen}{11.7\%} & \textcolor{red}{-4.0\%} \\
\rowcolor{gray!12}
\multicolumn{7}{l}{\textbf{Bounds}} \\
Uniform Guess & 61.2 & -- & \textcolor{red}{-16.1\%} & 0.203 & -- & \textcolor{red}{-16.7\%} \\
Empirical Best & -- & -- & -- & 0.125 & -- & \textcolor{darkgreen}{28.2\%} \\
\bottomrule
\end{tabular}
\caption{Comparison of model variants on accuracy and distribution distance metrics across unseen studies (\Cref{subsec:gen-unseen}). In each scenario, \protect\textbf{best scores are in boldface}, \protect\underline{second-best underlined}. Percent changes are relative.}
\label{tab:model-results}
\end{table*}


\myparagraph{Bounds on Metrics}
When computing the Wasserstein distance, we treat the responses in our dataset as the ground-truth distribution of human responses. Because individual responses naturally vary, this empirical sample may not perfectly capture the true distribution of outcomes under each condition. Our source experiments were highly powered to estimate a treatment effect of a certain size, not to robustly estimate the full distribution. To estimate an empirical upper bound on performance given this variance, we perform bootstrapping: we generate 100 resampled datasets (with replacement) from the original responses. For each bootstrapped dataset, we compute the Wasserstein distance between that resample and the full original sample, then average these distances across all 100 iterations. If our models Wasserstein distance meets or exceeds this, our error is no larger than the variability inherent in our data. We label this the ``\textbf{Empirical Best}''.

To establish a lower bound, we compare the standardized distribution of our predicted responses to a uniform distribution across $[0,1]$. We use this uniform baseline ``\textbf{Uniform Guess}'' to calculate a corresponding lower bound on both Wasserstein distance and accuracy.

\myparagraph{Comparing Metrics} Because the Wasserstein distance has a narrow range (e.g., 0.2 for a uniform guess, 0.1 for the empirical best), we report results as \emph{relative change} versus a baseline (either the base model with one-shot prompting or \texttt{GPT-4o} with one-shot prompting). For each method, we compute
$\frac{|a_{\mathrm{method}} - a_{\mathrm{base}}|}{|a_{\mathrm{base}}|} \times 100\%$,
assigning the sign so that positive values indicate improvement and negative values indicate regression. Here, $a_{\mathrm{method}}$ is the metric (e.g., Wasserstein distance) for the method under evaluation, and $a_{\mathrm{base}}$ is the metric for the baseline.

\myparagraph{Prompting Baselines} In our task formulation, we already incorporate all available demographic information from participants, which has proven an effective prompting mechanism in prior work \cite{hewitt2024predicting}. Prior work has also shown that reasoning over intermediate decisions \cite{wei2022chain} or using in-context-prompting \cite{dong2022survey} improves LLM prediction accuracy. Accordingly, we evaluate three baselines: (1) direct prediction prompting; (2) prompting to generate explicit reasoning traces before prediction; and (3) in-context prompting with few-shot examples. We select few-shot examples by finding the closest prompt stimuli neighbor via cosine similarity of embeddings, then choosing examples from five random participants. We use OpenAI's \texttt{text-embedding-3-large} model to embed stimuli. Prompt templates are provided in \Cref{sec:prompt} and \Cref{appendix:example-rtrace}.

\subsection{Generalization Across Unseen Studies}\label{subsec:gen-unseen}
First, we consider the case when there is \textbf{no in-domain studies data are available} to finetune on. Specifically, we assess the study-wise generalization of our finetuning methods by evaluating on studies that are completely out of domain (\emph{\emph{i.e.},}, not seen during training). 

\myparagraph{Setup} From the 210 studies in \dataname, we split 170 studies as our train-studies and 40 studies as test-studies. We train over 100\% of the training studies data, and evaluate on the 40 test studies. We compare the performance of our three finetuning objectives (SFT, SFT+Reasoning, and Contrastive DPO (\Cref{sec:finetuning})) across two open-source models, to three prompting baselines on \texttt{GPT-4o}. 

\myparagraph{Results} \Cref{tab:model-results} details the results on our evaluation metrics (\Cref{sec:eval_metrics}).  We find that finetuning meaningfully generalizes to unseen studies and improves the distributional alignment metric. Relative to the \texttt{GPT-4o} baseline, after SFT, \texttt{LLaMA3-8B} outperforms \texttt{GPT-4o} by \textbf{12.1\%}, and \texttt{Qwen2.5-14B} outperforms GPT4o by \textbf{13.2\%}, reflecting relative gains of 30.1\% and 26.3\% from fine-tuning. 
In open-source models, prompting through reasoning achieves significant gains, though our fine-tuned model still outperforms this baseline on distribution alignment. The best distributional alignment score (0.151) is achieved by fine-tuning \texttt{Qwen2.5-14B}; given an empirical best bound of 0.125 on this metric, our model \textbf{closely approximate actual human response distributions}.

Interestingly, although distributional alignment improves for unseen studies, response-level accuracy does not necessarily increase. This suggests that individual predictions may become less precise as we better approximate the distribution of how responses should look (for example, if user responses under condition follow a distribution $N(0,1)$, then a model that more accurately captures that distribution can incur higher error than one that always predicts the mean).  We include an in-depth discussion analyzing this in \cref{appendix:accuracy-vs-alignment}.

For predicting individual accuracy metric, contrastive DPO outperforms all other methods achieving 73.9\% accuracy. This is potentially due to our demographic-focused contrastive pair construction (\Cref{sec:finetuning}) that enables models to learn detailed distinction in simulating individual decisions, leading to more precise predictions of each individual.

\subsection{Generalization to Unseen Conditions and Outcomes} \label{sec:gen_cond_outcome}

Researchers often have topic-specific dataset they can leverage for finetuning. For example, a political polarization researcher may have existing data on how an intervention shifts feelings toward the opposing party and want to run studies to predict either (a) how this intervention influences a different outcome, such as respondents' confidence in government or (b) how a new intervention will affect that same outcome. Thus, in this section, we examine: \textit{How does finetuning help generalization to unseen conditions/outcomes within the same study?}

\myparagraph{Setup} We subset \dataname to all studies that contain at least 4 conditions / outcomes. We split the dataset into two splits: one for testing condition generalization, and one for testing outcome generalization. For the condition split, we randomly selected 75\% of the conditions across studies, and take all questions under those conditions for train, and hold out the other 25\% for test. For outcome splits, we repeat the same process but sampling on outcomes. For each, we finetune on the train set, then run our evaluation on the test set. \texttt{LLaMA3-8B-Instruct} is used as an example base model.

\begin{table}[t]
  \centering
  \footnotesize                               
  \setlength{\tabcolsep}{1pt}               
  \newcolumntype{Y}{>{\centering\arraybackslash}p{1.3cm}} 
  \begin{tabular}{@{}lYYYYY@{}}       
    \toprule
    \textbf{Model Variant} &
    \multicolumn{2}{c}{\textbf{Accuracy}$\uparrow$} &
    \multicolumn{2}{c}{\textbf{Distribution}$\downarrow$} \\
    \cmidrule(lr){2-3} \cmidrule(lr){4-5}
    & \textbf{Score} & \textbf{ vs Base} & \textbf{Score} & \textbf{ vs Base} \\
    \midrule
    \rowcolor{gray!12}\multicolumn{5}{@{}l@{}}{\textbf{Condition Split}}\\
    \texttt{LLaMA3-8B} Base                 & 71.0 & -- & 0.219 & --\\
    \quad + SFT                             & \textbf{74.2} & \textcolor{darkgreen}{4.5\%} & \underline{0.077} & \textcolor{darkgreen}{64.8\%} \\
    \quad + SFT w/ R.                & \underline{71.9} & \textcolor{darkgreen}{1.3\%} & \textbf{0.063} & \textcolor{darkgreen}{71.2\%} \\
    \quad + DPO                             & 71.2 & \textcolor{darkgreen}{0.3\%} & 0.208 & \textcolor{darkgreen}{5.0\%} \\
    Uniform Guess                           & 62.1 & -- & 0.180 & -- \\
    Empirical Best                          & --   & -- & 0.090 & -- \\  
    \midrule
    \rowcolor{gray!12}\multicolumn{5}{@{}l@{}}{\textbf{Outcome Split}}\\
    \texttt{LLaMA3-8B} Base                 & \underline{71.7} & -- & 0.224 & -- \\
    \quad + SFT                             & \underline{71.7} & \textcolor{red}{0.0\%} & \underline{0.125} & \textcolor{darkgreen}{44.2\%} \\
    \quad + SFT w/ R.                & 69.9 & \textcolor{red}{-2.5\%} & \textbf{0.114} & \textcolor{darkgreen}{49.0\%} \\
    \quad + DPO                             & \textbf{72.6} & \textcolor{darkgreen}{1.3\%} & 0.225 & \textcolor{red}{-0.5\%} \\
    Uniform Guess                           & 63.3 & -- & 0.165 & -- \\
    Empirical Best                          & -- & -- & 0.086 & -- \\
    \bottomrule
  \end{tabular}
  \caption{Performance metrics of \protect\texttt{LLaMA3-8B} under different training configurations, evaluated on 75\% train / 25\% held-out splits for both Outcome and Condition scenarios (\Cref{sec:gen_cond_outcome}). In each scenario, \protect\textbf{best scores are in boldface}, \protect\underline{second-best underlined}. Percent changes are relative.}
  \label{tab:condition-out-splits}
\end{table}

\myparagraph{Results} As shown in \Cref{tab:condition-out-splits}, when predicting held-out \textit{\textbf{conditions}}, our finetuned model with reasoning improves by \textbf{71\%} in estimating response distributions relative to the base model. Notably the distributional alignment after finetuning \textbf{surpass that of the ``Empirical Best'' threshold}, suggesting that the alignment with predictions in response to predictions is as close as another sample would be. SFT also increases the accuracy on individual predictions from 71.0\% to 74.2\%.

Across \textit{\textbf{outcomes}}, finetuning with reasoning also improves distribution distance, leading to 49\% relative improvement compared to \texttt{LLaMA3-8B} base. Generalization across unseen conditions tends to be greater than across unseen outcomes. This may be because LLMs grasp the underlying effects of how condition manipulations influence responses, but are more prone to misestimate the initial distribution of outcome questions. Thus, by holding the outcome constant and varying only the stimuli, the model can more effectively learn the resulting effects. This has especially practical value since studies often test many different condition stimuli on the same outcome (\emph{e.g.}, \citet{strand2024megastudy} tests 25 interventions to reduce partisan animosity).

\subsection{Generalization to Unseen Participants} \label{subsec:gen-participant}
Researchers often run pilot experiments on a small set of participants before committing to fully powered studies. Specific to such use case, we consider finetuning directly on a subset of participants and testing generalization to unseen participants in the same study. We examine how little data from a pilot study is needed to accurately predict outcomes for the remaining high-powered sample.

\myparagraph{Setup} We reuse the same study-level train-test split from \Cref{subsec:gen-unseen}: 170 studies for training and 40 for testing. 
For the 170 training studies, we randomly divide the participants \textbf{once} into: \textit{participant-train} (50\% of all individuals for training); \textit{participant-eval} (the remaining 50\% for evaluation of unseen participants \emph{but} seen studies).
From the \textit{participant-train} pool we draw progressively larger \emph{pilot} subsets corresponding to  
\(1, 5, 10, 20, 30, 40,50\%\) of \emph{all} participants (\emph{i.e.}, the \(50\%\) split \textit{all} of \textit{participant-train}).
For each split size we finetune the model and report performance on  
(i) the participant-eval splits of the \(170\) training studies (unseen individuals \emph{but} seen studies) and (ii) the \(40\) completely held-out studies (unseen individuals \emph{and} unseen studies).

This design reveals how much participant data a pilot must collect to obtain reliable generalization across both new participants and new studies.

\begin{figure}[t]
    \centering
\includegraphics[width=1\linewidth]{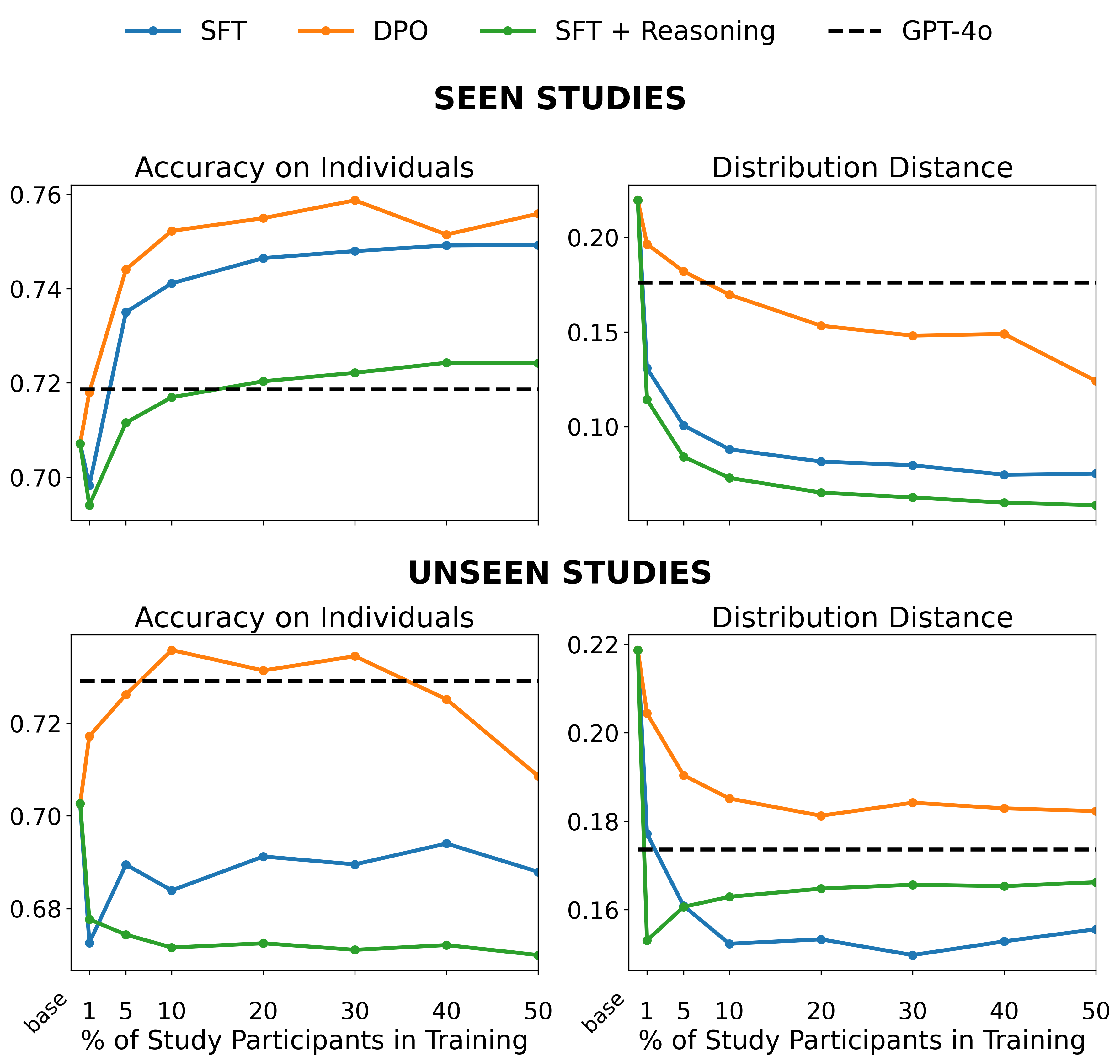}
    \caption{Learning curves on how \% of training samples generalizes to held-out participants in seen studies  and all participants in unseen studies, across varying participant size across studies (\Cref{subsec:gen-participant}). }
    \label{fig:seen-sweep}
\end{figure}

\myparagraph{Results} \Cref{fig:seen-sweep} shows the results on both the participants remained in the observed studies and all the participants in the observed studies. 

In seen studies (the 50\% held-out evaluation set), contrastive DPO tuning outperforms simple SFT for learning individual responses: with just 10\% of the data, accuracy rises from 71\% to 75\% (a ~13\% relative error reduction). However, SFT estimates the distribution more effectively. When augmented by reasoning traces, SFT+Reasoning brings even further alignment on the distribution. 
Across all studies, the learning curves in \Cref{fig:seen-sweep} indicate that saturation of learning is consistently achieved at around \textbf{10\%} of participant data. 

In unseen studies (the completely held-out 40 studies), a similar pattern holds. Contrastive DPO tuning also shows superior performance on the individual response accuracy, surpassing GPT4o with only 10\% of data. SFT yields better distribution alignment, whereas reasoning trace augmentation provides little benefit in this setting.

\begin{figure}[t]
\centering
\includegraphics[width=1\linewidth]{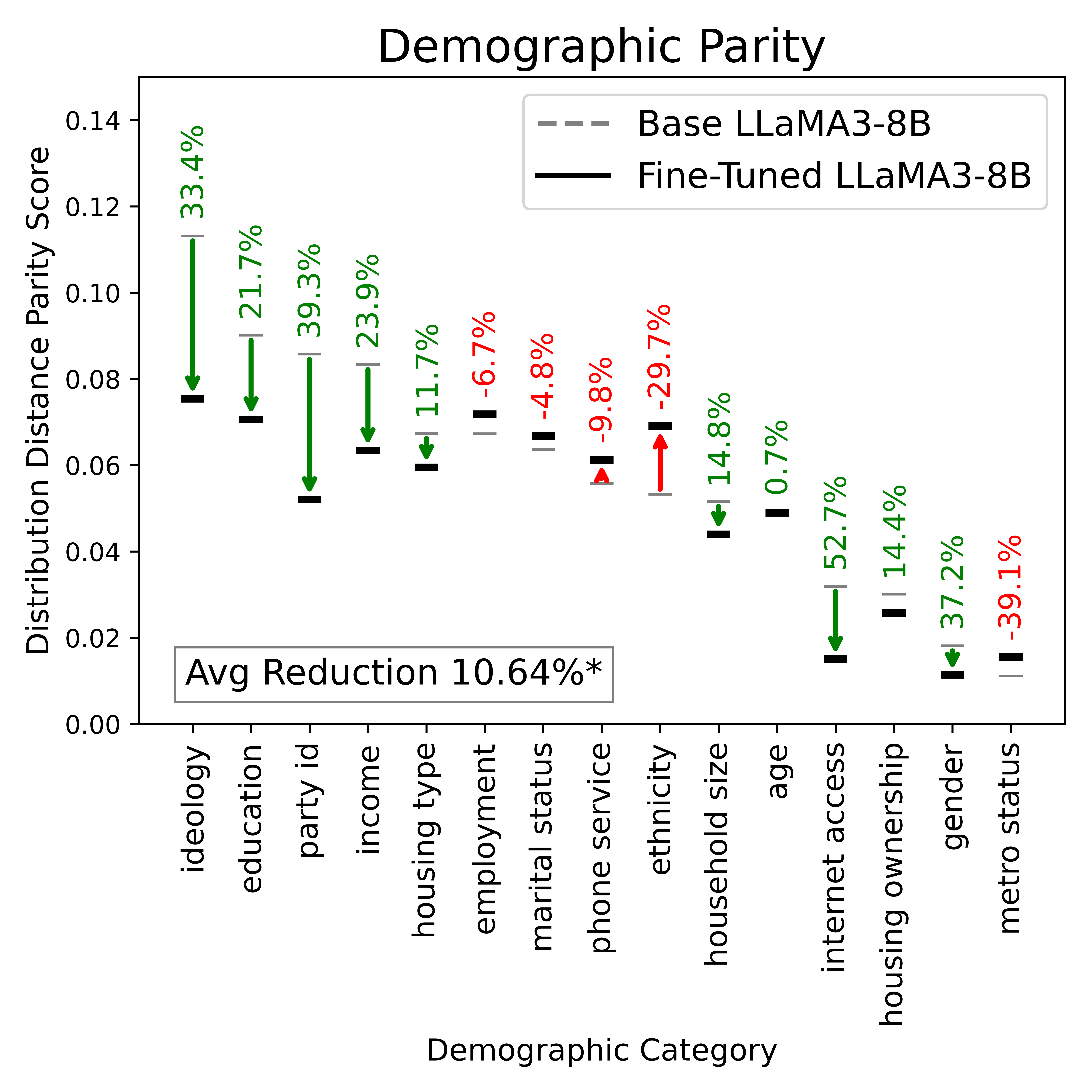}
\caption{Parity difference reduction in predicting distributions across demographic categories after finetuning \texttt{LLaMA-8B} (\Cref{sec:demo_bias})}
\label{fig:parity-reduce}
\end{figure}

\subsection{Demographic Bias and Parity} \label{sec:demo_bias}

A principal strength of our dataset is that it captures rich demographic attributes of all participants and how they may affect responses. In this section, we analyze the response distributions for the full-data SFT of \texttt{LLaMA3-8B}, described in \Cref{subsec:gen-unseen}. We further break down our distributional alignment metric (\emph{i.e.}, the Wasserstein distance between model responses under each unique condition and outcome) by demographic category: we subset the responses to each demographic subgroup and compute the average Wasserstein distance. Using this, we also compute demographic parity difference, a measure of bias \cite{jiang2022generalized} defined as the absolute gap between the highest- and lowest-performing demographic subgroups. 

After finetuning -- which improves overall distribution alignment -- we observe an average relative improvement of 28.5\% in distributional alignment across demographic categories (a full breakdown of improvement across every demographic subgroup is available in \Cref{appendix:demo-subsets}). More notably, across all demographic categories, parity difference is reduced by 10.6\%—a meaningful decrease in model bias, as shown in \Cref{fig:parity-reduce}.

\section{Conclusion} \label{sec:conclusion}
In this paper, we finetune LLMs to create a general use behavioral model that can accurately predict how individuals respond in social science experiments. We introduce \dataname, a standardized, large-scale dataset comprising 2.9 million individual responses from more than 400,000 participants across 210 social science experiments. Through fine-tuning \texttt{Qwen2.5-14B}, we create \modelnameQW, which relative to \texttt{GPT-4o}, produces predictions that are 13\% more distributionally aligned with real human responses.
Given the strong generalization we observe, we recommend that researchers begin finetuning on their existing datasets to yield more accurate and useful simulations. To support this, we open-source our dataset (\dataname), models (\modelname) and finetuning code. These provide a foundation in creating a unified behavioral prediction engine that can power simulations across every discipline.

\section*{Limitations} \label{sec:limitations}

Our participant sweeps in \Cref{subsec:gen-participant} show that performance tends to quick plateau with more participants' data. This is likely because we use relatively small parameter models. Given the size of our dataset, we anticipate that scaling to larger models (such as \texttt{Llama-70B} or \texttt{Llama-405B}) could further improve performance.

We rely on \texttt{GPT-4o-mini} to generate our oracle reasoning traces, and show that SFT on this traces does not always help improve performance for our task. Future work might investigate distilling those traces from more powerful reasoning models (\emph{e.g.}, OpenAI's \texttt{o3}), which could create more performant models.
Our models are trained exclusively on \dataname. Although our dataset is diverse, it contains only representative samples of the U.S. population for closed-form questions. We do not evaluate the generalization of our training to non-U.S. populations or to open-form questions. Future work should focus on integrating datasets from prior research into our training paradigm, constructing new datasets, and exploring training on open-ended responses, which may further enhance performance and generalization.

\section*{Ethics Statement} \label{sec:ethics}
We publicly release all data and our fine-tuned models. All materials were collected and processed in accordance with the respective data, checkpoint, and API usage policies. The dataset used in this study is drawn from publicly available, peer-reviewed social science experiments from the NSF’s TESS repository, all of which comply with established ethical and privacy standards. Our dataset includes stimuli that may be considered contentious, and our fine-tuned models may generate incorrect or unsafe content. While fine-tuning has led to meaningful improvements in model accuracy, it may also lead users to become overconfident in the results. We strongly advise all users to verify outputs carefully before deploying this work in real-world applications.

\section*{Acknowledgments}
We thank Omar Shaikh for helpful discussions on finetuning for this task. We acknowledge funding support from the Stanford Institute for Human-Centered Artificial Intelligence, Google, AXA, SCB X, Hanwha, and American Express.



\bibliography{MAIN}

\appendix

\section{Data Reconstruction Agent} \label{sec:data_reconstruction}

\begin{figure}[t]
    \centering
    \includegraphics[width=1\linewidth]{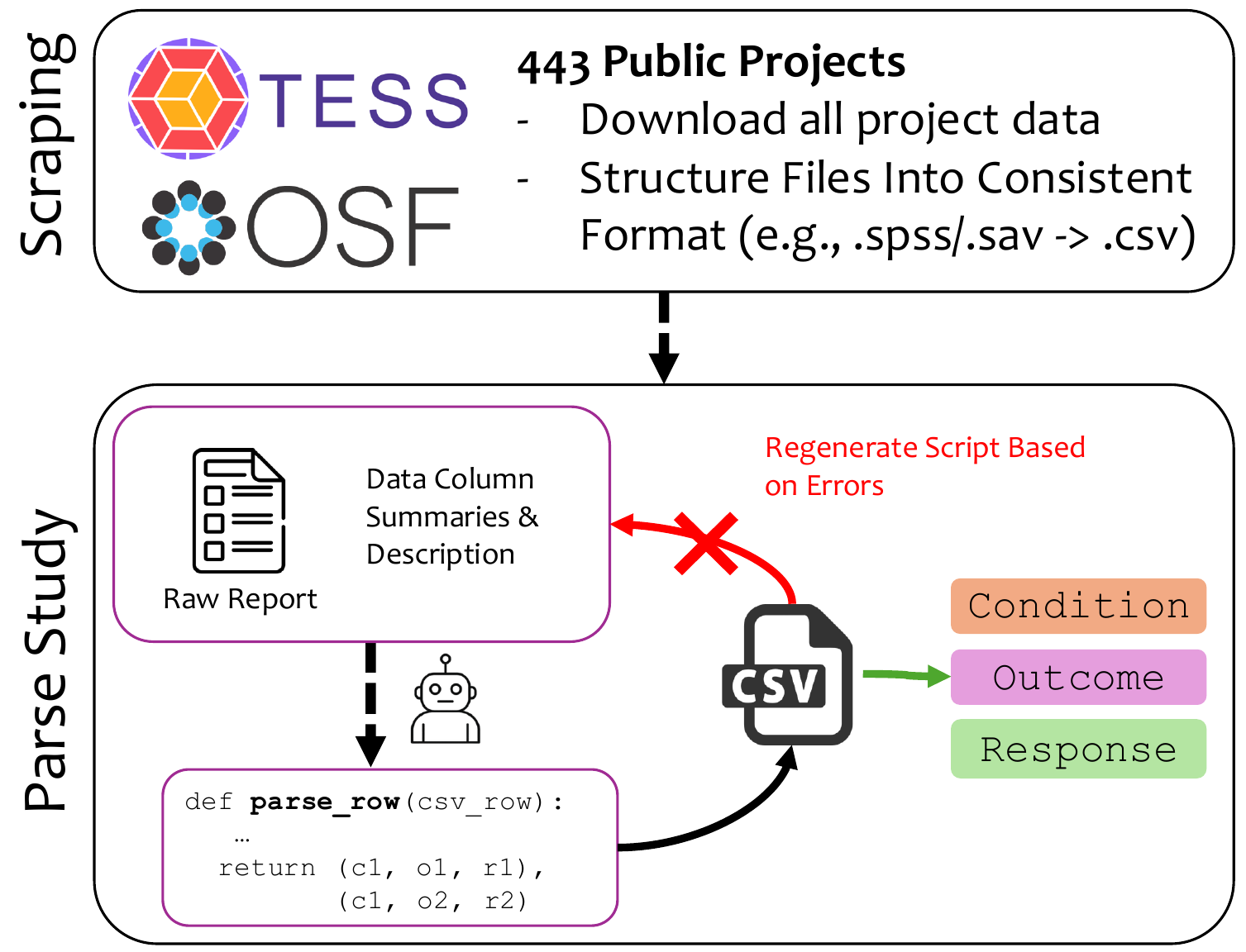}
    \caption{The workflow of our data construction agent for curating \dataname. The agent generates a script based on all study context and iteratively regenerates it for successful parsing of the data files from the source.}
    \label{fig:data-construction-workflow}
\end{figure}

In this section, we overview the workflow of our data construction agent. We intentionally avoided manual intervention to preserve a fully automated reconstruction pipeline.
As shown in \Cref{fig:data-construction-workflow}, we first download and standardize the publicly available data files form TESS \footnote{Data available via \href{https://osf.io/4547c/}{Open Science Framework repository}}, (\emph{e.g.}, converting data files to standard \texttt{.csvs}, and all \texttt{pdfs}/\texttt{docx} to text). Then we feed our data reconstruction agent with each research paper's full context (including description, data files, codebook, and stimuli). The agent then:
	1)	Identifies all experimental conditions;
	2)	Identifies the outcome questions;
	3)	Writes and executes parsing code to merge and clean the repository's CSV files, reconstructing each participant’s record, and combining the experimental condition and outcome question into a stimuli that maps to each participant response. This step goes through a generate-and-test cycle, which iterates until the code can reconstruct the dataset. 

    At the time of scraping, we pulled 443 publicly available project on the TESS OSF account. OSF projects often have a nested structure—where sub-projects can inflate this count. After deduplicating these, we identified only 321 unique studies, of which our agent successfully reconstructed 210. The following conditions were present in our actual reconstruction script: (1) must be able to find a description of the stimuli present for each condition (2) must reconstruct original/binary questions (3) outcome questions correspond to a condition-specific stimuli. If these aren’t satisfied the script will skip the study. A bulk of the failure cases were also just the inability of the LLM agent (powered by o4-mini-high) to accurately generate code that could parse the data into natural language, given the long context of the input data. A ``successful'' scrapes means the agent explicitly verifies two things during the code test-and-verify cycle: (a) The parsing code successfully compiles and executes without rasing errors (b) The parsing code generates non-empty outputs when operated row by row on the dataset.

\section{Discussion of Metric Evaluation: Accuracy vs. Distributional Alignment}
\label{appendix:accuracy-vs-alignment}

As we discuss in the main text, for our considered experiments, we value accurately modeling the Wasserstein distribution under experimental conditions above achieving high individual-level accuracy. In this appendix, we flesh out the intuition behind this. In treatment effect experiments, verifying the hypothesis involves simply a t-test comparing group means, relying solely on distribution-level attributes (mean and standard deviation) of responses to each condition. Individual responses, since they are parameterized purely based on the stimuli and demographic information, are inherently stochastic. 

Formally, the limitation of measuring individual response accuracy is that each response $r$ is parameterized only by a $(P, c, o)$ tuple, where $P$ consists of demographic keys. However, for a given demographic $P$ responding to a question $o$ after seeing stimulus $c$, we cannot confidently assert that the output should always be $r$. Instead, our training data merely shows the example $r$, drawn from the distribution $F(P, c, o)$, where $F$ is the underlying model of human responses. An ideal evaluation of a predictive model $F'$ would therefore compare the distribution $F'(P, c, o)$ to $F(P, c, o)$. However, we do not have enough data for each unique demographic to recover such an exact distribution $F(P, c, o)$. Instead, we compare $F(\cdot, c, o)$ to $F'(\cdot, c, o)$. Furthermore, even if a model were predicting the exact distribution $F(P, c, o)$, such an approach often performs poorly when reduced to single-example accuracy measurements.

As a simple example of this, consider our training data might contain ten 42-year-old males responding to the same stimulus: five individuals answer "0," two answer "1," and three answer "2." A perfect model representing this scenario would predict responses "0," "1," and "2" with probabilities of 50\%, 20\%, and 30\%, respectively. Yet, such predictions yield low accuracy scores when applied on our training data. Conversely, consistently predicting the midpoint ("1") improves accuracy but poorly represents the true distribution. 

Our intuition is that \texttt{GPT-4o} achieves higher accuracy because its responses often cluster narrowly around mid-scale values (e.g., in our evaluation set the standardized \texttt{GPT-4o} predictions have $\sigma=0.154$, while our \modelnameLA predictions have $\sigma=0.195$ and human responses $\sigma=0.192$ where $\sigma$ is sample standard deviation). 

To further illustrate the limitations of accuracy, we also compare two predictors: a) always predicting the question response scale mean vs. b) randomly sampling responses from the ground truth distribution of responses in the experimental condition. Across our 40 evaluation studies, the accuracy of method (a) outperforms method (b) for 30\% of cases, despite (a) being meaningless and (b) representing a model predicting perfectly.

\section{Implementation Details} \label{sec:implement_details}

\myparagraph{Finetuning Configurations}
In all finetuning experiments, we finetune models for $1$ epoch with a global batch size of $256$ on $8$ \textsc{NVIDIA-A100x80G} GPUs.
The learning rate (LR) is set to $1e{-05}$ for SFT and $1e{-06}$ for DPO. We adopt cosine LR scheduler with a warm-up ratio of $0.05$ and weight decay of $0.1$. Training times varied between experiments, but took roughly between 4 and 24 hours. During inference with all open-source models, we uniformly set temperature=$0.6$, top\_p=$0.9$, max\_length=$4096$. 

\myparagraph{Prompting Configurations}
When prompting OpenAI's proprietary models (e.g. \texttt{GPT-4o} for simulation, \texttt{GPT-4o-mini} for reasoning trace generation, \texttt{o4-mini-high} for dataset consutrction), we use default API parameters (\emph{e.g.}, temperature=$1$, top\_p=$1$). All metrics reported reflect results from a single experiment run (except when noted otherwise).

\newpage
\newpage

\section{Prompt Templates For Prediction} 

\label{sec:prompt}

\begin{tcolorbox}[colback=black!5!white,colframe=black!75!black,title=Prompt for Direct Prediction]
\textbf{[SYSTEM]}: 
You are simulating a survey respondent. Answer exactly as instructed, following the specified response format without additional commentary.
\\
\\
\textbf{[USER]} You are a survey respondent with the following demographic profile: \\
\texttt{{\color{red}\{\text{Demographic Info.}\}}}
\\
\\
Read the question below and answer exactly as this person would. Follow the response instructions precisely.
\\
\texttt{{\color{red}\{\text{Stimuli}\}}}
\end{tcolorbox}

\begin{tcolorbox}[colback=black!5!white,colframe=black!75!black,title=Example User Message of Direct Prediction]
\textbf{[USER]} 
You are a survey respondent with the following demographic profile: \\
- Age: 29 \\
- Gender: Female \\
- Education: Vocational/tech school/some college/associates \\
- Employment: Employed as paid employee \\
- Marital Status: Never married \\
- Housing Ownership: Occupied without payment of cash rent \\
- Household Size: 6 \\
- Ideology: Somewhat Liberal \\
- Phone Service: Cellphone only
\\
\\
Read the question below and answer exactly as this person would. Follow the response instructions precisely.
\\
\\
You read 'Emily recently graduated from high school and will attend college in the fall. Her mother and father, both factory workers, are very proud of her. Emily is excited to be attending her first-choice college, a highly-ranked private university. The university provides funding to cover the costs that families cannot pay, so Emily will graduate with no debt.' and then were asked: 'How unlikely or likely would you be to recommend history?' Only return an integer from 1 to 6, nothing else.
\end{tcolorbox}

\begin{tcolorbox}[colback=black!5!white,colframe=black!75!black,title=Modified System Message For Reasoning-Based Prompting]
\textbf{[SYSTEM]}: \\
You are simulating a survey respondent. You are to answer exactly as instructed, but also include your reasoning (5 sentences or less) before you output your answer.Please follow the exact output format below.
\\
\#\#\# Output format \\
<trace> \\
…your step-by-step reasoning here… \\
</trace> \\
PREDICTION: <verbatim answer> (conclude with predicted answer, use exactly the option label/number with no extra commentary)
\end{tcolorbox}

\begin{tcolorbox}[colback=black!5!white,colframe=black!75!black,title=Modified System Message for Few-Shot Prompting]
\textbf{[SYSTEM]}: \\
You are simulating a survey respondent. Answer exactly as instructed, following the specified response format without additional commentary.\\
\\
As you answer, consider how the following similar question was answered by other participants:\\
\\
Question: \texttt{{\color{red}\{\text{Sampled Similar Stimuli}\}}} \\
\\
Person 1 Profile: \\
\texttt{{\color{red}\{\text{Person 1 Demographic Info.}\}}} \\
Answer: \texttt{{\color{red}\{\text{Person 1 Answer}\}}} \\
\\
Person 2 Profile: \\
\texttt{{\color{red}\{\text{Person 2 Demographic Info.}\}}} \\
Answer: \texttt{{\color{red}\{\text{Person 2 Answer}\}}} \\
\\
\texttt{{\color{blue}\text{......}}}\\
\texttt{{\color{blue}\{\text{...More Persona Examples...}\}}}\\
\texttt{{\color{blue}\text{......}}}\\
\\
Person 5 Profile: \\
\texttt{{\color{red}\{\text{Person 5 Demographic Info.}\}}} \\
Answer: \texttt{{\color{red}\{\text{Person 5 Answer}\}}} \\
\end{tcolorbox}


\section{Reasoning Trace Generation Prompts}

We use \texttt{GPT-4o-mini} in order to generate "oracle" reasoning traces from our finetuning data. The model generates what an ideal reasoning trace \textit{should look like} given a prompt, response pair. 

\begin{tcolorbox}[colback=black!5!white,colframe=black!75!black,title=Prompt for Oracle Reasoning Traces]
\textbf{[SYSTEM]}:You are an expert behavioral scientist asked to write a plausible, forward‑looking reasoning trace that *predicts* which answer a survey respondent will give. Draw on knowledge of behavioral and social science theory to explain how and why this person responded the way they did. 
\\
\\
**Key constraints for the reasoning trace** \\
1. **Prospective viewpoint.** Write as if you do *not* know the final choice yet. Describe the mental steps a typical person with the given persona might take when first seeing the stimuli.
\\
2. **No answer leakage inside the trace.** The true answer is supplied only for your private verification. Do **not** quote, paraphrase, or rely on it within the narrative.
\\
3. Be concise but specific in your reasoning and avoid repetition. Keep the reasoning trace 5 sentences or less."
\\
\\
\#\#\# Output format
\\
<trace> …your step‑by‑step reasoning here (written as if before 'knowing' the answer)… </trace> \\
PREDICTION: <verbatim answer> (conclude with predicted answer, use exactly the option label/number with no extra commentary)
\\
\\
\textbf{[USER]} \\
**Persona**: \texttt{{\color{red}\{\text{Demographic Info.}\}}}
\\
**Stimuli**: \texttt{{\color{red}\{\text{Stimuli}\}}}
\\
<!-- TRUE ANSWER (use only to verify your prediction; do NOT reference inside <trace>): \texttt{{\color{red}\{\text{Ground-Truth Response}\}}} -->
\\
\\
Write the reasoning trace and final prediction now, following the format above.
\end{tcolorbox}

\label{appendix:example-rtrace}


\begin{tcolorbox}[colback=black!5!white,colframe=black!75!black,title=Example Prompt and Response with Reasoning Trace]
\textbf{Prompt:} You are a survey respondent with the following demographic profile:  \texttt{{\color{red}\{\text{Demographic Info.}\}}}
\\

Read the question below and answer exactly as this person would. Follow the response instructions precisely.
\\

You viewed an image and description of the Hyundai Kona Electric, a 100\% electric version of the Hyundai Kona that costs about \$40,000, has a 250 mile range and can accelerate from zero to 60 mph in about 6.4 seconds and then were asked: "What is your first reaction to the product?" Only return an integer from 1 to 5, where 1 means Very negative and 5 means Very positive, nothing else.
\\

\textbf{Reasoning:} Upon seeing the Hyundai Kona Electric, the individual may first consider their personal values and preferences, which lean strongly towards traditional vehicles and fossil fuels, aligning with their very conservative ideology. The relatively high price point of \$40,000 is likely to evoke concerns about affordability, especially given their household income bracket. Furthermore, the electric vehicle's technological aspects and environmental emphasis may clash with their skepticism towards climate change narratives and push towards eco-friendly solutions. Additionally, a family of four may prioritize practicality and utility over innovation, leading them to view a fully electric vehicle as less suitable for their needs. Overall, these factors suggest a negative response to the product.

\textbf{Response:} 1
\end{tcolorbox}

\section{Demographic Subset Results} 
\label{appendix:demo-subsets}
See \Cref{tab:category_improvement} for a breakdown of distribution distance under each demographic category of our dataset.

\begin{table*}[t]
\centering
\fontsize{8pt}{9.6pt}\selectfont
 \renewcommand{\arraystretch}{0.8}  
  \setlength{\extrarowheight}{-30pt}
\begin{tabular}{llrrr}
\toprule
Attribute & Category & LLaMA-8B & Finetuned & (\%) Improvement \\
\midrule
age & 65-100 & 0.2353 & 0.1679 & 28.63 \\
age & 35-50 & 0.2284 & 0.1571 & 31.23 \\
age & 50-65 & 0.2340 & 0.1599 & 31.66 \\
age & 25-35 & 0.2190 & 0.1484 & 32.23 \\
age & 0-25 & 0.2196 & 0.1510 & 31.24 \\
gender & Female & 0.1910 & 0.1342 & 29.72 \\
gender & Male & 0.1814 & 0.1165 & 35.80 \\
education & Post grad study/professional degree & 0.2381 & 0.1838 & 22.84 \\
education & Vocational/tech school/some college/associates & 0.2358 & 0.1612 & 31.64 \\
education & Bachelor's degree & 0.2339 & 0.1745 & 25.39 \\
education & High school graduate or equivalent & 0.2356 & 0.1527 & 35.18 \\
education & Some high school (no diploma) & 0.2602 & 0.1676 & 35.58 \\
education & Less than high school & 0.2578 & 0.1861 & 27.79 \\
employment & Self-employed & 0.1984 & 0.1506 & 24.12 \\
employment & Employed as paid employee & 0.1876 & 0.1380 & 26.46 \\
employment & Disabled & 0.2070 & 0.1565 & 24.38 \\
employment & Retired & 0.2006 & 0.1450 & 27.73 \\
employment & Looking for work & 0.1925 & 0.1437 & 25.37 \\
employment & Not working for other reasons & 0.1950 & 0.1424 & 26.97 \\
employment & Temporarily laid off & 0.2122 & 0.1740 & 18.01 \\
marital status & Divorced & 0.2412 & 0.1769 & 26.68 \\
marital status & Married & 0.2342 & 0.1652 & 29.45 \\
marital status & Never married & 0.2227 & 0.1528 & 31.40 \\
marital status & Living with partner & 0.2244 & 0.1581 & 29.52 \\
marital status & Widowed & 0.2441 & 0.1882 & 22.92 \\
marital status & Separated & 0.2507 & 0.2047 & 18.33 \\
housing ownership & Owned or being bought by you/someone in your household & 0.1909 & 0.1343 & 29.64 \\
housing ownership & Rented for cash & 0.1860 & 0.1300 & 30.09 \\
housing ownership & Occupied without payment of cash rent & 0.2116 & 0.1670 & 21.05 \\
housing type & A one-family house detached from any other house & 0.1906 & 0.1329 & 30.25 \\
housing type & A one-family house attached to one or more houses & 0.1847 & 0.1354 & 26.70 \\
housing type & A mobile home or trailer & 0.2092 & 0.1564 & 25.26 \\
housing type & Boat, RV, van, etc & 0.2151 & 0.1795 & 16.57 \\
housing type & A building with 2 or more apartments & 0.1695 & 0.0824 & 51.37 \\
metro status & Metro Area & 0.1876 & 0.1299 & 30.73 \\
metro status & Non-Metro Area & 0.1963 & 0.1451 & 26.10 \\
income & 50-74K & 0.2282 & 0.1560 & 31.65 \\
income & 40-49K & 0.2340 & 0.1645 & 29.70 \\
income & 20-29K & 0.2349 & 0.1482 & 36.89 \\
income & 200K+ & 0.2016 & 0.1648 & 18.25 \\
income & 125-149K & 0.2373 & 0.1708 & 28.03 \\
income & 75-99K & 0.2328 & 0.1671 & 28.22 \\
income & 30-39K & 0.2341 & 0.1584 & 32.31 \\
income & 100-124K & 0.2393 & 0.1761 & 26.43 \\
income & 150-175K+ & 0.2536 & 0.1962 & 22.66 \\
income & 175-200K+ & 0.2484 & 0.1936 & 22.05 \\
income & 15-19K & 0.1677 & 0.0563 & 66.45 \\
income & 10-14K & 0.2158 & 0.0892 & 58.66 \\
income & 5-9K & 0.1914 & 0.0857 & 55.25 \\
income & <5K & 0.1306 & 0.0395 & 69.78 \\
internet access & Internet Household & 0.1888 & 0.1316 & 30.28 \\
internet access & Non-internet household & 0.1932 & 0.1386 & 28.29 \\
household size & 0-3 & 0.2310 & 0.1574 & 31.86 \\
household size & 3-6 & 0.2266 & 0.1536 & 32.21 \\
household size & 6-9 & 0.2311 & 0.1644 & 28.83 \\
household size & 9-20 & 0.2851 & 0.2190 & 23.18 \\
phone service & Cellphone only & 0.1863 & 0.1312 & 29.59 \\
phone service & Have a landline, but mostly use cellphone & 0.1977 & 0.1377 & 30.36 \\
phone service & Have cellphone, but mostly use landline & 0.1965 & 0.1447 & 26.37 \\
phone service & Landline telephone only & 0.2130 & 0.1607 & 24.55 \\
phone service & No telephone service & 0.2232 & 0.1889 & 15.40 \\
party id & Moderate Democrat & 0.1784 & 0.1400 & 21.54 \\
party id & Don't Lean/Independent/None & 0.1822 & 0.1391 & 23.64 \\
party id & Strong Democrat & 0.1776 & 0.1554 & 12.48 \\
party id & Lean Republican & 0.1914 & 0.1456 & 23.89 \\
party id & Lean Democrat & 0.1705 & 0.1557 & 8.66 \\
party id & Strong Republican & 0.1855 & 0.1505 & 18.83 \\
party id & Moderate Republican & 0.1858 & 0.1466 & 21.10 \\
ideology & Somewhat Liberal & 0.1804 & 0.1514 & 16.06 \\
ideology & Moderate & 0.1878 & 0.1414 & 24.73 \\
ideology & Liberal & 0.1877 & 0.1357 & 27.68 \\
ideology & Somewhat Conservative & 0.2063 & 0.1568 & 23.98 \\
ideology & Conservative & 0.2093 & 0.1591 & 24.00 \\
ideology & Extremely Liberal & 0.2002 & 0.1494 & 25.36 \\
ideology & Declined to Answer & 0.2379 & 0.1592 & 33.08 \\
ideology & Extremely Conservative & 0.2342 & 0.1949 & 16.80 \\
ideology & Very Conservative & 0.1888 & 0.1649 & 12.66 \\
ideology & Very Liberal & 0.1809 & 0.1613 & 10.82 \\
ethnicity & Hispanic & 0.2642 & 0.1792 & 32.17 \\
ethnicity & White & 0.2679 & 0.1808 & 32.52 \\
ethnicity & 2+ Race & 0.2849 & 0.2044 & 28.24 \\
ethnicity & Other & 0.2652 & 0.1895 & 28.54 \\
ethnicity & Black & 0.2749 & 0.1850 & 32.70 \\
\bottomrule
\end{tabular}
\caption{Prediction improvement in Wasserstein distance under each demographic category.}
\label{tab:category_improvement}

\end{table*}

\end{document}